\def\year{2020}\relax
\documentclass[letterpaper]{article} 
\usepackage{aaai20}  
\usepackage{times}  
\usepackage{helvet} 
\usepackage{courier}  
\usepackage[hyphens]{url}  
\usepackage{graphicx} 
\usepackage{graphicx}  
\newcommand{\citet}[1]{\citeauthor{#1} \shortcite{#1}}
\newcommand{\citep}{\cite}

\usepackage[mathscr]{euscript}  
\usepackage{booktabs} 
\usepackage{amssymb}
\usepackage{amsmath}
\usepackage{todonotes}
\usepackage{multirow}

\newcommand{\tabref}[1]{Table~\ref{#1}}

\newcommand{\secref}[1]{Section~\ref{#1}}
\newcommand{\equref}[1]{Eq.~(\ref{#1})}
\newcommand{\F}{\boldsymbol{F}}
\newcommand{\f}{\boldsymbol{f}}
\newcommand{\g}{\boldsymbol{g}}
\newcommand{\G}{\boldsymbol{G}}
\newcommand{\kvec}{\boldsymbol{k}}

\newcommand{\h}{\boldsymbol{h}}
\newcommand{\x}{\boldsymbol{x}}
\newcommand{\y}{\boldsymbol{y}}
\renewcommand{\r}{\boldsymbol{r}}
\newcommand{\z}{\boldsymbol{z}}

\newcommand{\fRNN}{\f_{\text{\sc{RNN}}}}
\newcommand{\fGRU}{\f_{\text{\sc{GRU}}}}
\newcommand{\fASRNN}{\f_{\text{\sc{ASRNN}}}}

\newcommand{\mbf}[1]{\mathbf{#1}}

\title{System Identification with Time-Aware Neural Sequence Models}
\author{Thomas Demeester\\
Internet Technology and Data Science Lab, Ghent University - imec\\
thomas.demeester@ugent.be
}
 
\begin{document}

\maketitle

\begin{abstract}
Established recurrent neural networks are well-suited to solve a wide variety of prediction tasks involving discrete sequences. However, they do not perform as well in the task of dynamical system identification, when dealing with observations from continuous variables that are unevenly sampled in time, for example due to missing observations. We show how such neural sequence models can be adapted to deal with variable step sizes in a natural way. 
In particular, we introduce a `time-aware' and stationary extension of existing models (including the Gated Recurrent Unit) that allows them to deal with unevenly sampled system observations by adapting to the observation times, while facilitating higher-order temporal behavior.
We discuss the properties and demonstrate the validity of the proposed approach, based on samples from two industrial input/output processes. 
\end{abstract}

\section{Introduction}\label{sec:introduction}

The field of system identification, aiming to build mathematical models of dynamical systems based on observed data, has been a large active research area for many years, with several specialized sub-fields \cite{Ljung2013}. 
Within this general field, 
the topic of research in this paper is the joint application of numerical methods developed to solve systems of differential equations, with established techniques from the field of artificial neural networks.  

On the one hand, neural networks provide a highly flexible tool to train unknown parameterized functions to fit available data \cite{Goodfellow2016}. In particular, recurrent neural networks (RNNs), and especially variants such as Long Short-Term Memory (LSTM) networks \cite{hochreiter_97} or Gated Recurrent Units (GRU) \cite{Cho2014_GRU}, have become important general-purpose tools for modeling discrete sequential data, for example in the area of natural language processing \cite{Young2017}. 
These models are however not naturally suited to deal with sampled time series where the interval between consecutive samples may not be constant over time. Yet, so-called \emph{unevenly spaced time series} occur often in practice, due to missing data after discarding invalid measurements, or because of variations in the sampling times \cite{Eckner2014}. For example, a patient's clinical variables may only be measured at irregular moments. Or, observations from systems in meteorology, economics, finance, or geology might only be possible at irregular points in time. 
Multivariate data consisting of individual time series with different sample rates are also naturally treated as unevenly spaced time series. 
\citet{Eckner2014} further summarizes a number of disadvantages of transforming unevenly spaced data through resampling into evenly spaced data.

On the other hand, numerical methods such as the Runge-Kutta schemes \cite{Butcher2016} lead to highly accurate solutions for dynamical systems with known differential equations, and are \emph{by construction} able to deal with varying time intervals. 

The main research question that we tackle is therefore the following. Given a set of dynamical system observations, how can we build a model that can make use of the full expressive power of general-purpose RNNs, but that naturally deals with unevenly spaced time series? 

It has already been pointed out that the Euler scheme for numerically solving differential equations bears similarities with artificial neural network models containing residual connections \cite{Weinan2017,Zhu2018,Chang2019}. Naively applying this numerical scheme to RNNs for time series modeling, would indeed allow correctly dealing with uneven time intervals. 
However, unlike traditional RNNs, such schemes are incremental in nature, due to the residual connections along the time dimension, and therefore ill-suited for modeling stationary time series.  We will show how this can be resolved, leading to models that take advantage of both the bounded character and expressive power of well-known RNNs, and the time-aware nature of numerical schemes for differential equations.

%
The paper makes the following contributions:
\begin{itemize}
    \item We introduce an extension of general-purpose RNNs to deal with unevenly spaced time series (\secref{subsec:time_rnns}), 
    and which can be used in higher-order numerical schemes (\secref{subsec:RK}). In particular, we propose a solution for the 
    so-called `unit root' problem related to incremental recurrent schemes, to make the proposed approach suited for modeling stationary dynamical systems.
    \item We introduce a time-aware and higher-order extension of the Gated Recurrent Unit and Antisymmetric RNN (\secref{subsec:GRU}).
    \item We provide insights into the introduced models with experiments on data from two industrial input/output systems (\secref{sec:experiments}). The code required to run the presented experiments is publicly available\footnote{\url{https://github.com/tdmeeste/TimeAwareRNN}}.           
\end{itemize}



\section{Related Work}\label{sec:relatedwork}

This work is situated in between the fields of system identification, numerical techniques for differential equations, and deep neural networks. We refer the reader to standard works in these areas \cite{Ljung2013,Butcher2016,Goodfellow2016}, and focus in the following paragraphs on specifically related works.

\subsubsection{System Identification with Neural Networks}
More than two decades ago, \citet{Wang1998} introduced `Runge Kutta Neural Networks', to predict trajectories of time-invariant ODEs with unknown right-hand side. Their core idea of combining neural networks and Runge-Kutta schemes still applies to the underlying work. 

Several authors have recently put forward techniques for learning to compose partial differential equations from data.
\citet{Rudy2017} presented an efficient method to select fitting nonlinear terms, from a library of potential candidate functions and their derivatives. 
\citet{Raissi2018} proposed to use Gaussian Processes for learning system parameters, to balance model complexity and data fitting. 
\citet{Raissi2018b} introduced another interesting approach whereby two neural network models are simultaneously trained to model the solution as well as the right-hand side of unknown partial differential equations.

\citet{Rudy2019} presented a new paradigm on system identification from noisy data. Rather than assuming ideal trajectories, they proposed simultaneously predicting the measurement noise on the training data, while learning the system dynamics. For the latter, they used multilayer feed-forward neural networks in a Runge-Kutta scheme, which allows dealing with non-uniform timesteps.  They provided results on well-known autonomous dynamical systems including the chaotic Lorenz system and double pendulum.  
We focus on the use of general-purpose recurrent neural networks, which allow modeling input/output systems. Their ideas on predicting the measurement noise can however be applied in our setting, which provides an interesting future research direction. 

\citet{Ayed2019} proposed a general model for continuous state-space dynamics, based on the adjoint state method, and with the explicit Euler scheme for stepping through time. As in most of the related work mentioned above, they focus on initial-value problems, while we target input/output systems. 
Importantly, their model is able to capture the entire state dynamics even if these are only partly observed. In our work, we also consider systems for which the entire state is not captured by the observations alone, an ideal setting for RNNs which maintain a hidden state through time.

The work by \citet{Zhu2017} is related to our work in the endeavor to extend RNN models with temporal information. 
They propose the Time-LSTM, engineered for applications in the context of recommendation engines. Its so-called `time gates' are designed to model intervals over time, to capture users' short- and long-term interests.

The use of RNNs in modeling time series with missing data was explored by \citet{Che2018}. More in particular, they leveraged `informative missingness', the fact that patterns of missing values are often correlated with the target labels. They introduced the GRU-D, an extension of the GRU with a mask indicating which input variables are missing, as well as as a decay mechanism to model the fading influence of missing variables over time.

\subsubsection{Recent Advances in Machine Learning}
The successful use of neural networks in the field of dynamical systems, has in turn inspired some recent developments in the area of deep learning. 

\citet{Weinan2017} shared ideas on using dynamical systems to model high-dimensional nonlinear functions in machine learning, pointing out the link between the Euler scheme and deep residual networks. 

\citet{Zhu2018} started from the same link, in the area of image recognition, and suggested to extend the residual building blocks by multi-stage blocks, whereby suitable coefficients are learned jointly with the model (rather than relying on existing Runge-Kutta schemes). 

Recently, \citet{Chen2018_nips} introduced \emph{Neural ODEs}, a new family of neural networks where the output is computed using a black-box ODE solver (which could involve explicit Runge-Kutta methods, or even more stable implicit methods). They provide proof-of-concept experiments on a range of applications. These include the `continuous depth' setting, where ODE solvers (including a Runge-Kutta integrator) are shown to be able to replace residual blocks for a simple image classification task. They also discuss the `continuous time' setting, and propose a generative time-series model. That model represents the considered time series, which may be unevenly spaced in time, as latent trajectories.
These are initialized by encoding the observed time series using an RNN, and can be generated by the ODE solver at arbitrary points forwards or backwards in time. Training of the generative model is done as a variational auto-encoder. 
The complementary value of our work is in the fact that we provide an input/output dynamical systems formulation, and we believe that our extension from discrete RNNs to time-aware RNNs for dynamical systems makes it straightforward to apply. 

\citet{Chang2019} introduced the Antisymmetric RNN (henceforth written ASRNN), again based on knowledge of the Euler scheme for dynamical systems.  The use of an anti-symmetric hidden-to-hidden weight matrix in the incremental recurrent scheme (as well as a small diffusion term), ensured stability of the system.   
They demonstrated their model's ability to capture long-term dependencies, for a number of image recognition tasks cast as sequence classification problems.  What's more, the ASRNN model is naturally suited for modeling unevenly sampled time series: while Chang et al.~view the step size as a fixed  hyperparameter to be tuned during model selection, it can just as well be used as the actual time step, which we will further describe in \secref{subsec:GRU}. We will adapt the ASRNN to become better suited for modeling stationary systems. 

Finally, we want to point out that our use of the term `higher-order' models, refers to the order of the corresponding Runge-Kutta method. The term `higher order neural networks' has also been used for neural network layers with multiplicative, rather than additive, combinations of features at the input  \cite{Zhang2012}. Also, `higher order recurrent neural networks' may refer to RNNs where the current hidden state has access to multiple previous hidden states, rather than only the last one.

\section{Time-Aware RNNs}
\label{sec:theory}

After a general treatment on adapting recurrent neural networks to deal with unevenly spaced data (\secref{subsec:time_rnns}), we show how to use such models in higher-order Runge-Kutta schemes (\secref{subsec:RK}), and introduce the higher-order time-aware extensions for the GRU and ASRNN (\secref{subsec:GRU}).

\subsection{RNNs and ODEs} 
\label{subsec:time_rnns}

An RNN that models a discrete sequence of $N$ inputs $\{\x_{n}\}_{n=1}^{N}$ and outputs $\{\y_{n}\}_{n=1}^{N}$, can generally be described as follows
\begin{equation}
\begin{aligned}
\h_{n} &= \fRNN(\x_{n}, \h_{n-1})\\
\y_{n} &= \g(\h_{n})
\end{aligned}
\label{eq:rnn}
\end{equation}
The quantity $\h_{n}$ is called the hidden state at time step $n$, and is obtained by combining the input $\x_{n}$ with the hidden state $\h_{n-1}$ from the previous time step through the cell function $\fRNN$, which contains the recurrent cell's trainable parameters. The initial hidden state $\h_0$ can be fixed to the zero vector, or trained with the model parameters. 
The output function $\mathbf{g}$, which converts the hidden state $\h_n$ to the output $\y_n$, is typically a basic neural network designed for classification or regression, depending on the type of output data. The nature of the recurrent network is determined by $\fRNN$. It could be for instance an Elman RNN \cite{Elman1990} or a GRU. Variations to \equref{eq:rnn} are possible, for example for an LSTM where an additional internal cell state is passed between time steps.

Now consider an unknown continuous-time dynamical system, with inputs $\x(t)$ and outputs $\y(t)$ at time $t$. These do not necessarily cover the entire state space, as argued in \cite{Ayed2019}. We therefore explicitly introduce the state variable $\h(t)$ which, when observed at one point in time, allows determining the future system behavior, provided the system equations and future inputs are known. 
Many dynamical systems in science and technology can be described by an $n$'th order ordinary differential equation (ODE) \cite{Coddington1955}. Assuming this holds for the considered system, its system equations can be reduced to a first-order ODE in the $n$-dimensional state variable $\h$
\begin{equation}
\begin{aligned}
\frac{d\h(t)}{dt} &= \F\big(\x(t), \h(t)\big)\\
\y(t) &= \G(\h(t))
\end{aligned}
\label{eq:ode}
\end{equation}
whereby $\mathbf{G}$ represents a mapping from the latent state space to the output space.  We assume time-invariant systems, i.e., the system equations only indirectly depend on the time, through the state $\h(t)$ and a potential source $\x(t)$. 

If we discretize time into a sequence $\{t_n\}_{n=0}^N$, with potentially irregular step sizes $\delta_n = (t_{n+1}-t_n)$, and make use of the differential form of the derivative over a single time step, we can discretize the system equations as
\begin{equation}
\begin{aligned}
\h_{n+1} &= \h_n + \delta_n \F\big(\x_n, \h_n\big)\\
\y_{n+1} &= \G(\h_{n+1})
\end{aligned}
\label{eq:ode_discr}
\end{equation}
in which $\h_n$ is shorthand for $\h(t_n)$ and similarly for the other quantities.

Because of the similarities between the ODE discretization formulation (\equref{eq:ode_discr}) and the standard RNN formulation (\equref{eq:rnn}), we will be able to use RNNs to model the system equations from the considered unknown system. However, there are two important differences between both formulations. 

\subsubsection{Predicting `current' vs.~`next' output$\;$}
First of all, where the RNN allows predicting the output $\y_n$ from the input $\x_n$ at the \emph{same} position in the sequence (combined with the previous state $\h_{n-1}$), 
\equref{eq:ode_discr} only allows predicting the output  $\y_{n+1}$ at the \emph{next} time step $t_{n+1}$, given the input and state at the current time step $t_n$. 
For evenly sampled data, shifting the entire output sequence over one position still allows training a model that predicts the output $\y_n$ at the same point in time as the input $\x_n$ while using a valid ODE discretization scheme. This is no longer possible for unevenly spaced data. However, our preliminary experiments on the considered datasets (see \secref{sec:experiments}) show very little difference in prediction effectiveness, if we train a model that predicts either $\y_{n+1}$ or $\y_n$ from $\x_n$ (in the evenly spaced case). Yet, it should be implemented with care in the case of irregular spacing, to avoid ending up with incorrect discretization schemes. 

\subsubsection{Stationarity$\;$}
Secondly, the trainable function $\F$ in \equref{eq:ode_discr} is only responsible for the `residual' part besides $\h_n$ when calculating $\h_{n+1}$. In other words, the term $\h_n$ acts as a skip connection. Such skip connections form a key element in deep residual neural networks which have been highly successful for image recognition tasks \cite{He2016}. However, in the time dimension they can have unwanted side effects. 
Naively extending a standard RNN with cell function $\fRNN$ to deal with uneven time steps according to the Euler scheme, yields the following update equation
\begin{equation}
	\h_{n+1} = \h_n + \delta_n\fRNN(\x_n, \h_n) \label{eq:naive}
\end{equation}
Consider for example an LSTM, its cell function bounded between $-1$ and $+1$. When starting from a zero initial state, the bounds of the state $\h_n$ after $n$ time steps become $\pm\sum_{\nu=0}^{n-1}\delta_\nu$, or approximately $\pm\,n\,\mu_\delta$, with $\mu_\delta$ the average step size.  
This does not necessarily lead to numerical problems, but a model with a potentially linearly growing state is ill-suited for modeling stationary time series. As will be demonstrated in \secref{sec:experiments}, it leads to sub-optimal results. 

A similar problem arises for the well-studied discrete-time stochastic process of first order defined by
\begin{equation}
\h_{n+1} = \h_n + \epsilon_n \label{eq:stochastic_process}
\end{equation}
with $\epsilon_n$ a serially uncorrelated zero-mean stochastic process with constant variance. 
The `unit root' is a solution to the process' characteristic equation. It leads to a trend in the mean, and hence a non-stationary process \cite{Guidolin2018}. 
However, the `differenced time series' $\Delta \h_{n+1} = \h_{n+1} - \h_n = \epsilon_n$ no longer has this unit root, and is stationary. 

Our update equation (\ref{eq:naive}) similarly suffers from the unit root problem. We can therefore apply the idea of the differenced time series, but at the same time need to adhere to the general scheme (\ref{eq:ode_discr}) to correctly deal with variable sample intervals.   We hence propose to use the following function $\F$ in system equation~(\ref{eq:ode_discr})
\begin{equation}
\F\big(\x, \h\big) = \frac{1}{\mu_{\delta}} \big(\fRNN(\x, \h) - \h\big)  \label{eq:F} 
\end{equation}
with $\mu_\delta$ again denoting the average step size.
The first-order update equation for the state can then be written as
\begin{equation*}
\h_{n+1} = \big(1-\frac{\delta_n}{\mu_\delta}\big)\h_n + \frac{\delta_n}{\mu_\delta} \fRNN(\x_n, \h_n)
\label{eq:update_state_stable}
\end{equation*}
The unit root introduced by the term $\h_n$ in \equref{eq:naive} has indeed disappeared because the expected value of $(1 - \delta_n/\mu_\delta)$ is zero. Note that unlike $\epsilon_n$ in \equref{eq:stochastic_process}, the RNN cell function in \equref{eq:naive} is no zero-mean stochastic process, nor are its consecutive outputs uncorrelated. There is therefore no theoretical guarantee that the resulting model will be stationary. However, the neural network no longer has to explicitly `learn' how to compensate for the unit root problem induced by the skip connection, and we hypothesize that it is therefore more naturally suited to deal with stationary data. For simplicity, in the remainder this will be called the `stationary' formulation, in contrast to the incremental models with the unit root, for simplicity denoted as the `non-stationary' one.

Combining equations (\ref{eq:ode_discr}) and (\ref{eq:F}) results in an \emph{extension} of the standard RNN scheme towards unevenly spaced time series, in the sense that for evenly spaced samples ($\delta_n=\mu_{\delta}$) they reduce to the standard RNN in \equref{eq:rnn},
or strictly speaking, one applied to predicting $\y_{n+1}$ from $\x_n$ and $\h_n$. 

Note that the sample rate at inference time might not entirely correspond to $\mu_\delta$, leading to a remnant of the unit root effect. Our experimentation indicates that this is less of a problem than for instance the presence of outliers (i.e., occasionally very large gaps between consecutive samples).

\subsection{Higher Order Neural Sequence Models}\label{subsec:RK}
A well-known family of higher order iterative ODE methods are the explicit Runge-Kutta methods \cite{Butcher2016}. Applying an $s$-stage Runge-Kutta method to the ODE in \equref{eq:ode} leads to the update equation
\begin{equation}
\h_{n+1} = \h_{n} + \delta_n \sum_{i=1}^{s} b_i\ \kvec_i \label{eq:RK}
\end{equation}
where $\kvec_1, \ldots, \kvec_s$ are found recursively over $s$ stages, as
\begin{align*}
\kvec_1 &= \F\big(\x_n, \h_{n}\big)\\
\kvec_i &= \F\Big(\x(t_n + c_i\delta_n), \h_{n} + \delta_n\sum_{j=1}^{i-1} a_{ij}\kvec_j\Big),\; {\scriptstyle i\in\{2,\ldots,s\} }
\end{align*}
for predefined values of the coefficients $b_i$, $c_i$, and $a_{ij}$. The error in predicting $\h_{n+1}$ from a correct $\h_{n}$ is called the \emph{local truncation error}. A Runge-Kutta method of order $p$ denotes a method where the local truncation error is of the order $\mathcal{O}(\delta^{p+1})$. 
The coefficients for a number of well-known methods are listed in \tabref{tab:RK}. For these methods, the number of stages equals their order.
The update scheme initially proposed in \equref{eq:ode_discr} corresponds to Euler's method ({\sc Euler}), the simplest Runge-Kutta method. 
We further consider the second-order Explicit Midpoint method (\textsc{Midpoint}), 
Kutta's third order method (\textsc{Kutta3}), and the classical fourth-order Runge-Kutta method (\textsc{RK4}).

\begin{table}[t]
	\caption{Non-zero coefficients of selected explicit Runge-Kutta methods.}
	\label{tab:RK}
		\centering
		\begin{sc}
		\resizebox{\linewidth}{!}{%
			\begin{tabular}{ l  @{\hspace{.2cm}}  c @{\hspace{.1cm}} c @{\hspace{.1cm}} c @{\hspace{.3cm}} c @{\hspace{.1cm}} c @{\hspace{.1cm}} c @{\hspace{.1cm}} c @{\hspace{.3cm}} c @{\hspace{.1cm}} c @{\hspace{.1cm}} c @{\hspace{.1cm}} c }
				\toprule
				Name (\textnormal{order}) & $c_2$ & $c_3$ & $c_4$ & $b_1$ & $b_2$ & $b_3$ & $b_4$ & $a_{21}$ & $a_{31}$ & $a_{32}$ & $a_{43}$ \\
				\midrule
				Euler (1)  &&&& 1 &&&&&&& \\
				Midpoint (2)  & 1/2 &&&& $1$ &&& 1/2 &&& \\
				Kutta3 (3)  & 1/2 &1&&1/6&2/3&1/6&&1/2&-1&2& \\
				RK4 (4)  & 1/2 &1/2&1&1/6&1/3&1/3&1/6&1/2&&1/2&1 \\
				\bottomrule
			\end{tabular}
		}
		\end{sc}
\end{table}

If the function $\F$ is differentiable, the model can be trained by backpropagating the gradient of the loss on the predicted output through the considered sequence, and within each time step, through the stages of the considered Runge-Kutta update scheme. 
Existing RNN models $\fRNN$ can hence be directly extended towards higher-order models that are suited for unevenly spaced data, by combining the proposed function $\F(\x, \h)$ in \equref{eq:F} with the Runge-Kutta update equation~(\ref{eq:RK}).

Some recent works have already applied numerical ODE methods in combination with neural networks to model dynamical systems \cite{Chen2018_nips,Rudy2019}. However, they do not investigate input/output systems, the focus of this work.  An important restriction to the use of higher-order methods for input/output systems, is the fact that the inputs $\x$ may be only available under the form of samples $\{\x(t_n)\}_{n=1}^N$. 
A valid Runge-Kutta update scheme however requires evaluating the input $\x(t_n+c_i\delta_n)$ at intermediate points in time (as per \equref{eq:RK}).  
As shown experimentally in \secref{sec:experiments}, this can be achieved through interpolation, but the approximation may counteract the effectiveness of higher-order methods. Another approach is to build a generative higher-order model for the input as well, which will be explored in future work.


\subsection{Time-Aware GRU and ASRNN} \label{subsec:GRU}

To make the results from the previous sections more tangible, 
we write out the proposed time-aware extension for the GRU cell and for the ASRNN. 

The cell function $\fGRU$ for the GRU \cite{Cho2014_GRU}, cast for the formulation $\h_{n+1} = \fGRU(\x_n,\h_n)$, is given by 
\begin{align*}
\fGRU  =& \; (1-\z_n)\odot\tanh{ \Bigl( W_h\x_{n}+U_h(\r_{n}\odot \h_{n})+b_h \Bigr) } \\
& + \z_n \odot \h_n
\end{align*}
in which $\odot$ represents elementwise multiplication. The auxiliary vectors $\z_{n}$ and $\r_{n}$ are called the `gates' 
\begin{align*}
\z_{n} &= \sigma(W_z\x_{n}+U_z\h_{n}+b_z)\\
\r_{n} &= \sigma(W_r\x_{n}+U_r\h_{n}+b_r)
\end{align*}
with $\sigma(.)$ the sigmoid function. 
With the hidden state dimension $k_h$ and input size $k_x$, the trainable parameters are given by the weight matrices $W_h, W_z, W_r \in \mathbb{R}^{k_h\times k_x}$, $U_h, U_z, U_r \in \mathbb{R}^{k_h\times k_h}$, and biases $b_h, b_z, b_r \in \mathbb{R}^{k_h}$. 
Applying \equref{eq:F} to avoid the unit root, yields for the time-aware GRU 
\begin{align*}
&\F(\x_n, \h_n) = \;\frac{1}{\mu_\delta}\bigl(\fGRU(\x_n, \h_n) - \h_n\bigr)\\
&\quad = \;\frac{\tilde\z_n}{\mu_{\delta}}  \odot  \Bigl(  \tanh \bigl(W_h\x_{n}+U_h(\r_{n}\odot \h_{n})+b_h\bigr)   - \h_n  \Bigr) 
\label{eq:HOGRU}
\end{align*}%
whereby for convenience $(1-\z_n)$ is replaced by a new gate $\tilde\z_n$.
The function $\F(\x_n, \h_n)$ is to be used with \equref{eq:ode_discr} for the first-order scheme, or with \equref{eq:RK} for its higher-order counterparts.  These equations retain the expressiveness and amount of trainable parameters from the original GRU cell, but remain valid for unevenly spaced data without inducing the unit root problem, and can be used in higher-order schemes.  In \secref{sec:experiments}, the non-stationary formulation from \equref{eq:naive}, i.e.,~$\F = \fGRU$, will be used as a baseline, to underline the importance of avoiding the unit root.

As a second example, we consider the gated ASRNN introduced by \citet{Chang2019}. 
Its cell function $\fASRNN$ can be written as
\begin{align*}
&\fASRNN = \; \z_n\odot\tanh\bigl( W_h\x_n + A\,\h_n + b_h \bigr) \\
&\text{with the gate}\;\; \z_n = \sigma\bigl( W_z\x_n + A\,\h_n + b_z \bigr)
\end{align*}
The hidden-to-hidden matrix $A\in\mathbb{R}^{k_h\times k_h}$ can be written as $A = (W_h - W_h^{T} - \gamma I)$. It corresponds to an anti-symmetric matrix (i.e., the difference between a weight matrix $W_h$ and its transpose $W_h^T$), with  a small negative value on the diagonal (indicated by the non-negative `diffusion' parameter $\gamma$ and unit matrix $I$) to ensure stability. 
The original ASRNN formulation follows the incremental (i.e., non-stationary) first-order formulation, with the step size $\epsilon$ as a hyperparameter for weighting the residual term in the state update equation. We replace it by the actual step size $\delta_n/\mu_\delta$ (normalized by its mean) to deal with uneven sample times, and keep the scaling factor $\epsilon$ for tuning the model:
\begin{equation}
\h_{n+1} = \h_{n} + \epsilon\,\frac{\delta_n}{\mu_\delta} \fASRNN(\x_n, \h_n) \label{eq:asrnn_non_stationary}
\end{equation} 
Its stationary counterpart follows from \equref{eq:ode_discr} and \equref{eq:F}
\begin{equation}
\h_{n+1} = \h_{n} + \frac{\delta_n}{\mu_\delta} \bigl(\epsilon\,\fASRNN(\x_n, \h_n) - \h_n\bigr) \label{eq:asrnn_stationary}
\end{equation} 
with straightforward extension to higher-order schemes.

\section{Experimental Validation}\label{sec:experiments}
This section presents experimental results on two input/output datasets. The main research questions we want to investigate are (i) what is the impact of unevenly sampled data on prediction results with standard RNNs vs.~with their time-aware extension, (ii) how important is the use of state update schemes that avoid the unit root issue, (iii) is there any impact from applying the time-aware models with higher-order schemes, and (iv) how are the output predictions affected while applying such higher-order schemes based only on \emph{sampled} inputs?

After describing the selected datasets (\secref{subsec:data}), the model architecture, and the training and evaluation setup~(\secref{subsec:setup}), we describe the experiments and discuss the results (\secref{subsec:results}).

\subsection{Datasets}\label{subsec:data}
We have selected two datasets from STADIUS's Identification Database ``DaISy'' \cite{Demoor1997}, a well-known database for system identification.

\paragraph{CSTR Dataset}
We use the \emph{Continuous Stirred Tank Reactor} (CSTR) dataset\footnote{\footnotesize{\url{ftp://ftp.esat.kuleuven.be/pub/SISTA/data/process_industry/cstr.dat.gz}}}. 
It contains evenly sampled observations (10 samples per minute) from a model of an exothermic reaction in a continuous stirred tank reactor. There is a single piecewise constant input signal, the coolant flow, and two output signals, the resulting concentration and temperature.  The data was first studied by \citet{Lightbody1997}, who introduced a basic neural network model in the context of adaptive control of nonlinear systems.  
It consists of a sequence of in total 7,500 samples, of which we used the first 70\% for training, the next 15\% for validation, and the last 15\% for testing. We used the original dataset for an evenly spaced baseline model, and generated a version with missing data, by randomly leaving out samples with a probability of $p_{\text{missing}} = 0.50$. The average interval between consecutive samples is doubled ($\mu_\delta=0.2$ minutes), but the data now contains gaps up to 13 times the original gap ($\delta_n\in[0.1, 1.3]$). 
We normalize the input and outputs, by subtracting their respective mean value in the training data, and dividing by the standard deviation.

\paragraph{Winding Dataset} 
We also use the data from a \emph{Test Setup of an Industrial Winding Process}\footnote{\url{ftp://ftp.esat.kuleuven.be/pub/SISTA/data/process_industry/winding.dat.gz}} (Winding). It contains a sequence of 2,500 evenly sampled measurements (10 samples per second). The test setup consists of 3 reels, the `unwinding reel' from which a web is unwinded, after which it goes over a traction reel, and is rewinded onto the `rewinding reel'.  
The inputs are the angular speed of the 3 reels, as well as the setpoint current of the motors for the unwinding and rewinding reel, i.e., five inputs in total. The two outputs correspond to measurements of the web's tension in between the reels.  The data was introduced and first studied by \citet{Bastogne1997}. We again created an artificial unevenly sampled data sequence based on real-world data, by randomly leaving out samples with probability $p_{\text{missing}} = 0.50$. We used the first 70\% of the input sequence for training, then 15\% for development, and the last 15\% for testing.

\subsection{Experimental Sfinaletup}\label{subsec:setup}

The overall goal of this work is to demonstrate how standard RNNs can be applied to unevenly sampled data from input/output models. To keep the approach generic, we use the same overall architecture and training procedure for both datasets. 

\paragraph{Model Architecture}
Modeling the data using a recurrent network with the same input dimension $k_{x}$ as the number of observed system inputs, appeared not sufficient. We therefore use RNN cell functions with a potentially higher input size $k$ and the same state size, and feed it with the observed input data (1 dimension for CSTR, 5 for Winding) extended to $k$ dimensions by applying a trainable linear mapping from the inputs to $k$ dimensions, followed by a $\tanh$ non-linearity.
As output function $\mathbf{G}(\h)$ (from \equref{eq:ode_discr}) we use a trainable linear mapping from $k$ state dimensions to the observed output space (2 dimensions for both datasets). 

\paragraph{Evaluation}
We report the root relative squared error (RRSE) averaged over the $k_\text{out}$ output channnels. Being a relative metric, it allows comparing results between different models and datasets. The RRSE is defined as \cite{Botchkarev2018}
\begin{equation*}
    \text{RRSE} = \frac{1}{k_\text{out}} \sum_{i=1}^{k_\text{out}} \sqrt{ \frac{ \sum_{n=1}^N\bigl(\hat{y}_n^{(i)}-y_n^{(i)}\bigr)^2 }{ \sum_{n=1}^N\bigl(\hat{y}_n^{(i)}-\mu_y^{(i)}\bigr)^2 } 
    }
\end{equation*}
in which $\hat{y}_n^{(i)}$ represents the predicted test value for channel~$i$ at time step $n$, whereas $y_n^{(i)}$ is the corresponding ground truth value, with $\mu_y^{(i)} = \frac{1}{N}\sum_n y_n^{(i)}$ the channel mean. 

Per output channel, the RRSE can be interpreted as the root mean squared (RMS) error of the prediction, normalized by the RMS error of the channel's average as a baseline. In other words, for an RRSE value of 100\%, the model performs no better than predicting the mean.

The test value is obtained by evaluating the model through the entire sequence, to ensure that the test sequence receives the appropriate initial state, and calculating the RRSE on the test sequence only. All reported values are the mean (and standard deviation) over five training runs starting from a different random initialization of the network.

\paragraph{Training}
We train the model with backpropagation-through-time \cite{Werbos1988} over 20 time steps, and perform the optimization in parallel over mini-batches of (possibly overlapping) segments of 20 time steps. The hidden state $\h_0$ at the start of the entire sequence is randomly initialized, and trained with the model. After each training epoch over all segments, a forward pass through the entire train sequence is performed, and the resulting states are used as initial state for the corresponding training segments (i.e., we used the training `scheme 4' as introduced by \citet{Deboom18_rnn}).  We minimize the mean squared error of the predicted outputs using the Adam optimizer \cite{Kingma2014}, and apply early stopping by measuring the RRSE on the validation sequence.

In order to investigate whether the proposed models work out of the box, rather than requiring substantial tuning, we only tune the baseline GRU and ASRNN models, without missing data. The same hyperparameters are then adopted for the experiments in the uneven sampling setting. They
are shown in \tabref{tab:hyperparams}. 

During our preliminary experiments, we noticed that applying regularization through dropout gave higher prediction errors. Given the small amount of training data (i.e., a single sequence of a few thousand measurements), our hypothesis is that applying dropout while allowing for larger numbers of trainable parameters did not allow to better capture the system dynamics, while it made training more difficult. We therefore chose to tune the model complexity only through the hidden state size $k$.

\begin{table}[t!]
	\caption{Hyperparameters tuned over ranges $b\in\{64, 512\}$, $k\in\{5,10,20,30,40,60,80,100,150\}$, and $\lambda \in\{0.001,0.003,0.01\}$. }
	\label{tab:hyperparams}
	    \centering
		\resizebox{.9\columnwidth}{!}{%
			\begin{tabular}{lcccc}
				\toprule
				\multirow{2}{*}{hyperparameter} & \multicolumn{2}{c}{CSTR} & \multicolumn{2}{c}{Winding} \\
				& GRU & ASRNN & GRU & ASRNN \\
                \midrule
                minibatch size $b$      & 512 & 512 & 512 & 64 \\
                state size $k$          & 20 & 100 & 10 & 10 \\
                learning rate $\lambda$ & 0.001 & 0.001 & 0.003 & 0.01 \\
				\bottomrule
			\end{tabular}
		}
\end{table}

\subsection{Results and Discussion}\label{subsec:results}

\subsubsection{Baselines}
The test error for the GRU and ASRNN baselines without missing data are shown in \tabref{tab:baselines} (top two lines). As mentioned above, the hyperparameters from \tabref{tab:hyperparams} are tuned on these. For both models, the formulation without unit root is used. For the GRU, this comes down to its standard formulation, whereas for the ASRNN, the stationary variant in \equref{eq:asrnn_stationary} is used, which for evenly sampled data reduces to
$
\h_{n+1} = \epsilon\,\fASRNN(\x_n, \h_n).
$

Overall it can be seen that the prediction error for the CSTR data is much lower than for the Winding data. This is likely due to properties of the datasets.  The CSTR dataset contains smooth simulation results, and has a relatively higher sample rate with respect to changes in the signal compared to the Winding data, which consists of actual measurements. 

We created a number of baselines for the data with missing samples as well, also shown in \tabref{tab:baselines}. When the standard GRU is applied, ignoring the missing data, there is a substantial increase of the error (`standard, ignore missing' in \tabref{tab:baselines}). The results are not dramatic, though. We hypothesize that the model learns, to some extent, to compensate in its output for sudden larger gaps in the input (due to larger temporal gaps). From that perspective, it does make sense to apply a standard RNN to unevenly sampled data. 
A straightforward way to augment standard RNN models with variable time steps, is by providing the step size $\delta_n$ (normalized) as an additional input signal to the model. This reduces the error by a few percentage points (`standard, extra input $\delta_n$' in the table). More advanced models, in line with \cite{Zhu2017}, may provide an even better alternative.

Finally, we applied the incremental Euler scheme of \equref{eq:naive} without compensation of the unit root, both for the GRU and the ASRNN (indicated as `time-aware, non-station.'). This leads to an increased error, confirming the hypothesis from \secref{subsec:time_rnns} that the presence of the unit root negatively affects the modeling of stationary data. Note however that the ASRNN seems less affected than the incremental GRU model.
The step size $\epsilon$ and the diffusion parameter $\gamma$ from the original ASRNN were both set to $1.0$ for the baseline without missing data, but were now tuned over the same values as in \cite{Chang2019}, without much improvement. 

\begin{table}[t!]
	\caption{Baseline results with missing data. Displaying test RRSE values in percentage points (mean $\pm$ std).}
	\label{tab:baselines}
		\centering
		\resizebox{\columnwidth}{!}{%
			\begin{tabular}{l   @{\hspace{.1cm}}  c  @{\hspace{.2cm}}  c}
				\toprule
				Model  & CSTR & Winding \\
                \midrule
				\multicolumn{3}{l}{\textit{Baselines on all original samples}} \\
                GRU (standard) & $\mbf{2.2 \pm 0.5}$ & $\mbf{20.1 \pm 0.6}$ \\
                ASRNN (stationary) & $2.5\pm 0.2$ & $27.4\pm 3.9$ \\  
                \midrule 
				\multicolumn{3}{l}{\textit{Baselines with missing data}} \\
                GRU (standard, ignore missing) & $10.8\pm 0.6$ & $30.2 \pm 0.4$ \\  
                GRU (standard, extra input $\delta_n$) & $\mbf{9.2\pm 1.4}$ & $\mbf{28.9\pm 1.7}$ \\
                GRU (time-aware, non-station.) & $79.7\pm 8.0$ & $64.8\pm 10.2$ \\
                ASRNN (time-aware, non-station.) & $12.3\pm 1.1$ & $41.5\pm 7.4$ \\
				\bottomrule
			\end{tabular}
		}
\end{table}

\subsubsection{Time-aware higher-order models}
The results for the stationary time-aware higher-order GRU model are shown in \tabref{tab:higherorder}.
The `constant' vs.~`linear' input interpolation shown in the table is related to the issue identified in \secref{subsec:RK} that correct higher-order schemes require inputs evaluated in between samples, i.e.,  $\x(t_n+c_i\delta_n)$, with $c_i$ depending on the specific Runge-Kutta scheme. For the CSTR data, the constant approximation $\x(t_n+c_i\delta_n)\approx\x_n$ is sufficient as the inputs are piecewise constant over several time steps, and higher-order schemes lead to lower errors. However, this is not the case for the Winding dataset, where some of the inputs correspond to continuous variables. The results in \tabref{tab:higherorder} indeed show that higher-order schemes are not beneficial when the inputs are assumed piecewise constant within each sample interval (column `constant' for the Winding data).
However, with a simple linear interpolation between consecutive inputs $\x(t_n+c_i\delta_n)\approx (1-c_i)\,\x_n + c_i\,\x_{n+1}$, it seems the output error again decreases for the tested higher-order schemes. More advanced interpolation methods may be more suited still, but were considered out of scope for this work.

For both datasets, the {\sc Euler} scheme performs a few percentage points worse than the standard GRU where the time steps are not explicitly encoded (see baseline `ignore missing' in  \tabref{tab:baselines}). 
This might be related to the absence of hyperparameter tuning for the time-aware models. However, increasing the order leads to gradually lower test errors for the time-aware methods with appropriate input interpolation. The 4'th order {\sc RK4} method leads to the overall lowest error for the missing data setting, even without tuning.  

Due to space constraints, \tabref{tab:higherorder} only shows time-aware results for the stationary GRU. Note that the corresponding ASRNN errors would remain slightly higher, consistent with the baselines. Also, the non-stationary counterpart of \tabref{tab:higherorder} would show that the presence of the unit root annihilates the positive effect of higher-order schemes entirely. 

\begin{table}[t!]
	\caption{Stationary time-aware higher-order GRU with constant vs.~linear input interpolation for the datasets with missing data. Displaying test RRSE values in percentage points (mean $\pm$ std).}
	\label{tab:higherorder}
		\centering
		\resizebox{0.9\columnwidth}{!}{%
			\begin{tabular}{l@{\hspace{.4cm}}c@{\hspace{.6cm}}cc}
				\toprule
				\multirow{2}{*}{Scheme} & CSTR & \multicolumn{2}{c}{Winding}  \\
				& constant & constant & linear \\
                \midrule
                \sc{Euler}  & $12.1\pm 1.3$ & $33.1\pm 0.6$ & $33.1\pm 0.6$ \\
                \sc{Midpoint}  & $11.0\pm 4.1$ & $35.3\pm 2.1$ & $28.2\pm 1.8$ \\ 
                \sc{Kutta3}  & $9.9\pm 4.7$ & $34.7\pm 4.3$ & $27.1\pm 1.0$ \\ 
                \sc{RK4}  & $\mbf{8.0\pm 0.4}$ & $32.2\pm 6.6$ & $\mbf{25.6\pm 1.4}$ \\ 
				\bottomrule
			\end{tabular}
			}
\end{table}

\subsubsection{Summary}
We now shortly look back at the research questions formulated at the start of this section.
In our setting, randomly leaving out training and test samples from a sequence of input/output system measurements leads to an increase in output prediction error. However, standard RNNs can still make meaningful predictions, especially when the temporal information is explicitly provided as a feature. For the datasets under study, the proposed time-aware higher-order schemes have the potential to compensate even better for the missing data. Eliminating the unit root appears however important when applying a standard RNN cell in an incremental update scheme. Furthermore, for data with continuously valued input samples, the use of higher-order schemes only makes sense if a proper interpolation in the input space is performed.

\section{Conclusions and Future Research Ideas}\label{sc:conclusions}

This paper focused on using neural sequence models for input/output system identification from unevenly spaced observations. We showed how to extend standard recurrent neural networks to naturally deal with unevenly spaced data by augmenting the update scheme with the local step size in a way that allows modeling stationary dynamical systems, and showed how the resulting model can be used in higher-order Runge-Kutta schemes.

We provided experimental results for two different input/output system datasets where we experimented with the impact of randomly leaving out data samples. Applying the time-aware model in higher-order schemes, gave better output predictions compared to ignoring the uneven sample times. The direct extension of RNNs with the incremental Euler scheme to correctly account for uneven sample times appeared to give inferior results. We hypothesized that this was due to the unit root problem, leading to an inherently non-stationary model, and showed how to avoid that problem. 

Future research includes looking into more complex input/output systems with non-uniform noisy data. 
One potential research direction could involve the application of adaptive sampling schemes during forecasting. 
For example, the introduced models could be readily used with the Runge-Kutta-Fehlberg methods with adaptive step size \cite{Fehlberg1969} in a computationally efficient way. A further promising research direction is in extending the proposed techniques for dynamical systems to generative sampling models for time series as proposed by \citet{Chen2018_nips}. 
A potentially interesting application domain would be in robotics, where light-weight dynamical system models with adaptive sample times could be of interest in terms of computational efficiency.

\section*{Acknowledgments}
This research received funding from the Flemish Government under the \emph{“Onderzoeksprogramma Artifici\"ele Intelligentie (AI) Vlaanderen”} programme. I am grateful to Cedric De Boom for his feedback on the paper, and to the anonymous reviewers for their useful suggestions.


\fontsize{9.0pt}{10.0pt} \selectfont 

\bibliography{aaai20}
\bibliographystyle{aaai}
\end{document}